\documentclass{article}
\usepackage{amssymb}
\usepackage{xcolor}

\usepackage[final]{corl_2025} 

\usepackage{bm}
\usepackage{url}
\usepackage{cite}
\usepackage{tikz}
\usepackage{xfrac}
\usepackage{xargs}
\usepackage{times}
\usepackage{array}
\usepackage{siunitx}
\usepackage{relsize}
\usepackage{amsmath}
\usepackage{wrapfig}
\usepackage{mathrsfs}
\usepackage{scalerel}
\usepackage{stfloats}
\usepackage{flushend}
\usepackage{multicol}
\usepackage{graphics}
\usepackage{graphicx}
\usepackage{hyperref}
\usepackage{multicol}
\usepackage{verbatim}
\usepackage{multirow}
\usepackage{booktabs}
\usepackage{pgfplots}
\usepackage{etoolbox}
\usepackage{algorithm}
\usepackage{xinttools}
\usepackage{glossaries}
\usepackage{algorithmic}
\usepackage[nolist]{acronym}
\usetikzlibrary{calc, backgrounds}
\pgfplotsset{compat=1.16}
\graphicspath{{images/}}
\DeclareGraphicsExtensions{.pdf,.jpeg,.png,.jpg}
\definecolor{wong_gray}{HTML}{888888}%
\definecolor{wong_black}{HTML}{333333}%
\definecolor{wong_gold}{HTML}{E69F00}%
\definecolor{wong_cyan}{HTML}{56B4E9}%
\definecolor{wong_green}{HTML}{009E73}%
\definecolor{wong_yellow}{HTML}{F0E442}%
\definecolor{wong_blue}{HTML}{0072B2}%
\definecolor{wong_red}{HTML}{D55E00}%
\definecolor{wong_pink}{HTML}{CC79A7}%
\definecolor{wong_magenta}{HTML}{CA1963}%
\definecolor{cool}{HTML}{3B4DBF}%
\definecolor{warm}{HTML}{B30326}%
\newacro{fov}[FOV]{Field of View}
\newacro{fcn}[FCN]{Fully Convolutional Network}
\newacro{ood}[OOD]{Out of Domain}
\newacro{rf}[RF]{Receptive Field}
\newacro{wsl}[WSL]{Weakly Supervised Learning}
\newacro{bce}[BCE]{Binary Cross Entropy}
\begin{document}
\title{%
%
Self-supervised Learning Of Visual Pose Estimation\\Without Pose Labels By Classifying LED States
}
\author{Nicholas Carlotti$^{1}$, Mirko Nava$^{1}$, and Alessandro Giusti$^{1}$%
 \thanks{$^{1}$All authors are with the Dalle Molle Institute for Artificial Intelligence (IDSIA), USI-SUPSI, Lugano, 6962, Switzerland \texttt{nicholas.carlotti@idsia.ch}. This work is supported by the Swiss National Science Foundation, grant number 213074.}%
}%
\maketitle
\begin{abstract}
We introduce a model for monocular RGB relative pose estimation of a ground robot that trains from scratch without pose labels nor prior knowledge about the robot's shape or appearance.
At training time, we assume: (i) a robot fitted with multiple LEDs, whose states are independent and known at each frame; (ii) knowledge of the approximate viewing direction of each LED; and (iii) availability of a calibration image with a known target distance, to address the ambiguity of monocular depth estimation.
Training data is collected by a pair of robots moving randomly without needing external infrastructure or human supervision.
Our model trains on the task of predicting from an image the state of each LED on the robot.
In doing so, it learns to predict the position of the robot in the image, its distance, and its relative bearing.
At inference time, the state of the LEDs is unknown, can be arbitrary, and does not affect the pose estimation performance.
Quantitative experiments indicate that our approach: is competitive with SoA approaches that require supervision from pose labels or a CAD model of the robot; generalizes to different domains; and handles multi-robot pose estimation.
\end{abstract}
\keywords{Self-supervised Learning, Pretext Task, Visual Pose Estimation}
%

\begin{figure}[th]
    \centering
    \includegraphics[height=28mm]{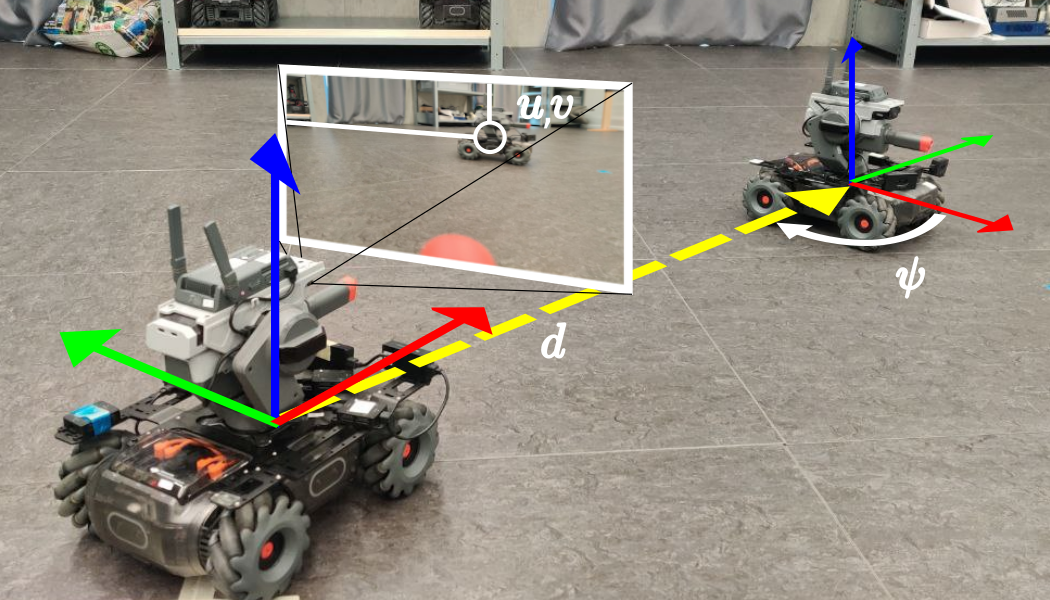}
    \includegraphics[height=28mm]{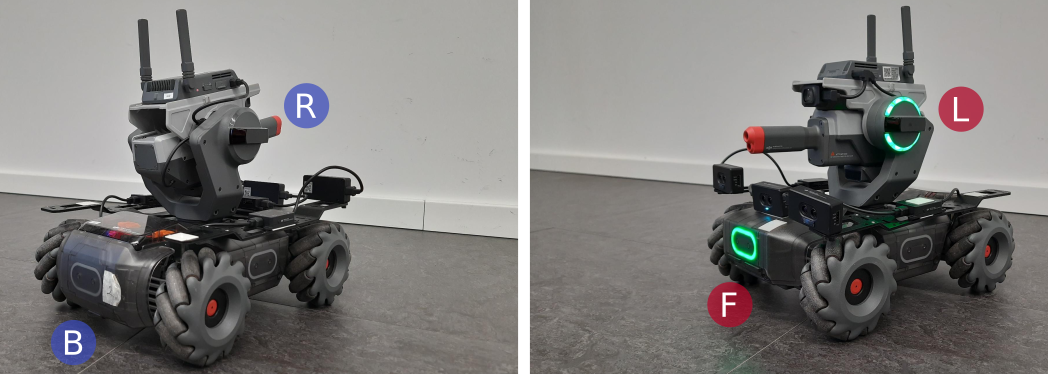}
    \caption{
    By solving the multi-LED state classification task ({\color{cool}blue} for off LEDs, {\color{warm}red} for on; F for front, B for back, L for left, R for right), our model learns from scratch to estimate the location of the robot $(u,v)$ in the image, its relative distance $d$, and relative bearing angle $\psi$ w.r.t. the camera. These variables are used in combination with the camera intrinsics to recover the pose of the robot.}
    \label{fig:intro}
\end{figure}

\section{Introduction}\label{intro}
Relative localization of mobile robots is fundamental for applications involving multiple robots that must coordinate with each other~\citep{dorigo2021swarm}. 
SoA pose estimation approaches solve the problem by training a deep neural network in a supervised way, assuming access to a large dataset of images representing the robot, each labeled with its true relative pose~\citep{wang2021gdr, liu2025gdrnpp}.
Acquiring such a dataset in real environments is expensive and requires external infrastructure for generating ground truth labels~\citep{jing2020self}.
Acquiring it in simulation requires a realistic and textured CAD model of the robot, while trained predictors suffer from the sim-to-real gap~\citep{salvato2021crossing}.
Novel object pose estimation approaches also assume access to CAD models, used to generate templates with known robot poses and matched with patches of the input image using learning-based descriptors~\citep{nguyen2023cnos, ausserlechner2023zs6d, moon2024genflow}.
However, the availability of visually-accurate and up-to-date CAD models of the robot is a strong requirement, especially for custom-built platforms.
%

In this work, we assume no access to data labeled with robot poses nor prior knowledge of the robot's 3D shape or texture.
In contrast, we assume each robot to be equipped with $K$ LEDs that can be independently turned on and off, and to know the approximate direction from which each LED is visible relative to the robot's heading.
In this challenging scenario, we propose a novel approach for training a robot pose estimation model from scratch:
training data is collected fully autonomously, without any external supervision, by a pair of robots that move in an arbitrary, possibly unknown way in the environment, without a shared frame of reference.
Each robot broadcasts (e.g., via a radio link) the true state (on or off) of all of its LEDs, which are toggled multiple times during data collection, independently from each other, and in a random way.
Our approach handles data that features a visible robot infrequently: considering that robot poses are unavailable during data collection, each robot will be visible in the other's \ac{fov} only in a small, unknown subset of the frames acquired.
Given a frame of the camera taken from a robot, called observer, our model predicts the state of each of the $K$ LEDs of the other robot, called target, i.e., a classification problem on $K$ independent binary labels. This is a \emph{pretext task} as solving it is not our ultimate goal; indeed, our \emph{end task} is to predict the full pose (position and orientation) of the target.

During inference, the LEDs are no longer needed: our model estimates, directly from a camera frame, the robot's position in the image, its bearing relative to the camera, and its apparent image size (see Figure~\ref{fig:intro}).
%
We compute the metric distance of the robot from its apparent image size with a calibration based on a single image depicting the robot and annotated with its distance from the camera, a similar assumption to Depth Anything~\citep{yang2024depth, yang2024depthv2}.

To the best of our knowledge, no other SoA approach learns pose estimation without pose labels or a CAD model of the robot.
Through a careful design of the neural network architecture and loss function, the model is forced to understand the robot's structure, which is crucial for pose estimation.
%
%
%
We further remark that the model is given no information on how the LEDs appear visually; our model learns to recognize them as part of its classification task.
The only assumption is that the LED state affects the robot's appearance in a way that is observable from an image of the robot acquired from a given, approximately known range of directions.

Our \textbf{main contribution} is methodological; we propose an approach for learning visual pose estimation of robots by training on the self-supervised pretext task of multi-LED state classification.
%
%
Experimental results indicate that:
(i) the approach trains pose estimation models that are competitive with SoA approaches requiring supervision from pose labels or a CAD model of the robot;
(ii) training is robust to data featuring a visible robot only in a limited amount of camera frames; 
(iii) the LED state does not impact the pose estimation performance; 
(iv) models generalize to unseen environments with no fine-tuning and are capable of multi-robot pose estimation;
(v) the approach can fine-tune a pre-trained model to a different deployment environment.

\section{Related Work}\label{sec:rw}

\paragraph{Supervision in Visual Object Pose Estimation.}\label{sec:rw:visual}
Approaches designed for visual object pose estimation can be directly applied to robots.
In this context, different assumptions about the object's appearance are made, serving as supervision during training.
Traditional approaches assume access to a large dataset labeled with object poses~\citep{xiang2018posecnn,wang2021gdr,liu2025gdrnpp}.
A less strict assumption is to have access to a realistic textured CAD model of the object, used to generate inexpensive simulated data~\citep{pham2022pencilnet, li2022self, deng2020self, leeCameratoRobotPoseEstimation2020,su2022zebrapose}.
Recent works leverage deep template matching~\citep{labbe2023megapose} to estimate the pose of novel objects, further leveraging foundation models such as Segment Anything Model (SAM)~\citep{kirillov2023segment} to segment objects that are matched using DINOv2~\citep{oquab2023dinov2} features with rendered templates of the object at known poses~\citep{nguyen2023cnos}.
The pose of the matched template is refined with a point matching stage~\citep{lin2024sam, nguyen2024gigapose}, with P$n$P~\citep{ausserlechner2023zs6d, ornek2024foundpose}, or by iteratively minimizing the optical flow between the template and segmented image~\citep{moon2024genflow}.
Approaches based on NeRF~\citep{mildenhall2020nerf} or Gaussian Splatting~\citep{kerbl3Dgaussians} assume access to images of a static scene labeled with the camera pose. They are used to estimate the pose of objects by iteratively minimizing the photometric error with~\citep{yen2020inerf} or without an initial pose estimate~\citep{bortolon2024iffnerf,matteo20246dgs}.

All the above methods learn pose estimation on data annotated by strong supervision sources (e.g., tracking system, textured CAD model of objects, calibrated views).
By contrast, our proposed approach makes no such assumption: we rely on pose-free, real-world data autonomously generated by two robots and labeled only with the binary state of multiple and independent LEDs.

\paragraph{Weakly Supervised Learning in Computer Vision.}\label{sec:rw:wsl}
In \ac{wsl}, object detection and image segmentation are learned by training on a classification task with inexpensive image-level labels~\citep{zhang2021weakly}.
Class Activation Map (CAM)~\citep{zhou2016learning,ramaswamy2020ablation, jiang2021layercam} approaches 
are used to find the most discriminative areas for classifying the image.
Crucially, discriminative areas for an image classification task depict the object of interest, enabling object detection~\citep{lu2020geometry} and segmentation~\citep{xie2022c2am}.
\ac{wsl} enables manipulation from image-level labels ~\citep{jang2018grasp2vec}, where the reward signal for an RL agent tasked to pick and place objects is derived from the embeddings of an image depicting an object and another with the object removed.
In our work, we take inspiration from \ac{wsl} in computer vision and go beyond a simple detection task: we introduce a novel approach for full pose estimation of robots, learned by the model as a result of solving a multi-label binary classification task.

\paragraph{Self-supervised Robot Learning.}\label{sec:rw:ssrl}
Learning complex robotic tasks from real-world data requires labels as a form of supervision, whose collection is time-consuming and expensive.
To reduce reliance on labeled data, approaches pre-train models on pretext tasks to learn valuable pattern recognition skills~\citep{radosavovic2023real, antsfeld2024self, qian20243d, nava2024self, carlotti2024learning}, and later fine-tune them with a limited amount of labels on the end task of interest:
as an example, approaches mask image patches and task a model to fill out the missing areas, given as input the masked image~\citep{radosavovic2023real} and additional views of the scene~\citep{antsfeld2024self, qian20243d}.
Recently, works showed that LED state classification pretext task is conducive to fine-tuning with labeled data on the task of visual robot detection~\citep{nava2024self}, and pose estimation~\citep{carlotti2024learning}.
By contrast, our approach does not require fine-tuning or strongly annotated data to learn robot pose estimation.

\section{Method}\label{sec:method}
Given an RGB image $\mathbf{I} \in \mathbb{R}^{W \times H \times 3}$ of width $W$ and height $H$  collected by the observer robot, our model estimates the 2D pose $\mathbf{P} = \langle x,y,\psi \rangle$ of the target robot relative to observer's camera.
Additionally, the model classifies the state (on or off) of $K$ independent LEDs mounted on the target's body and visible from a known range of directions.
Formally, we define the deep learning model $m\bm{_\theta}(\bm{I}) = \langle \hat{u}, \hat{v}, \hat{d}, \hat{\psi}, \hat{\bm{l}} \rangle$ where $\bm{\theta}$ are the model parameters; $\hat{u}$ and $\hat{v}$ are the image-space coordinates of the robot; $\hat{d}$ is the distance of the robot from the camera in the scene; $\hat{\psi}$ is the robot's orientation relative to the camera; and $\hat{\bm{l}} = \langle \hat{l}_1, \dots, \hat{l}_K \rangle$ are the predicted probabilities that each of the $K$ LEDs is turned on.
Using the model prediction, we recover the robot pose by back-projecting its image location using the camera intrinsic parameters, selecting the point at the estimated distance from the optical center, and combining it with the rotation $\hat{\psi}$.
%
We optimize the model parameters $\bm{\theta}$ through gradient descent with a \ac{bce} loss defined on the LED states; details on the neural network architecture and training hyper-parameters can be found in the appendix.

Our model architecture is a \ac{fcn}~\citep{long2015fully} composed solely of convolution and pooling layers.
It outputs a set of two-dimensional feature maps (maps for short) composed of cells.
%
Specifically, we exploit that each output cell of a \ac{fcn} attends only to a local area of the input image, represented by the \ac{rf}.
Given a monocular image, our model produces an LED state map $\hat{L}^k$ for each LED, a localization map $\hat{P}$, and a relative bearing map $\hat{\Lambda}$.
All these maps have the same $H' \times W'$ shape.
Each cell of the $\hat{L}^k$ map takes values in the $[0,1]$ range, indicating the confidence that the $k$-th LED is turned on or off (represented by 1 and 0, respectively).
If the LED is not visible inside the \ac{rf}, the cell will have a value of 0.5 to indicate uncertainty; this is the case for most cells whose RF captures background areas of the image.
To this end, we define a multi-label binary classification task on the LED states, with the loss
$\mathcal{L}_\text{led}^k = \operatorname{BCE}(\hat{l}^k, l^k)$
%
computed between each cell $\hat{l}^k$ of the LED state map $\hat{L}^k$ and the ground truth state $l^k$ of the $k$-th LED;
as such, the loss $\mathcal{L}_\text{led}^k$ is itself composed of maps divided into cells, having one map for each of the $K$ LEDs.

\begin{figure}[t]
    \centering
    \includegraphics[width=1\linewidth]{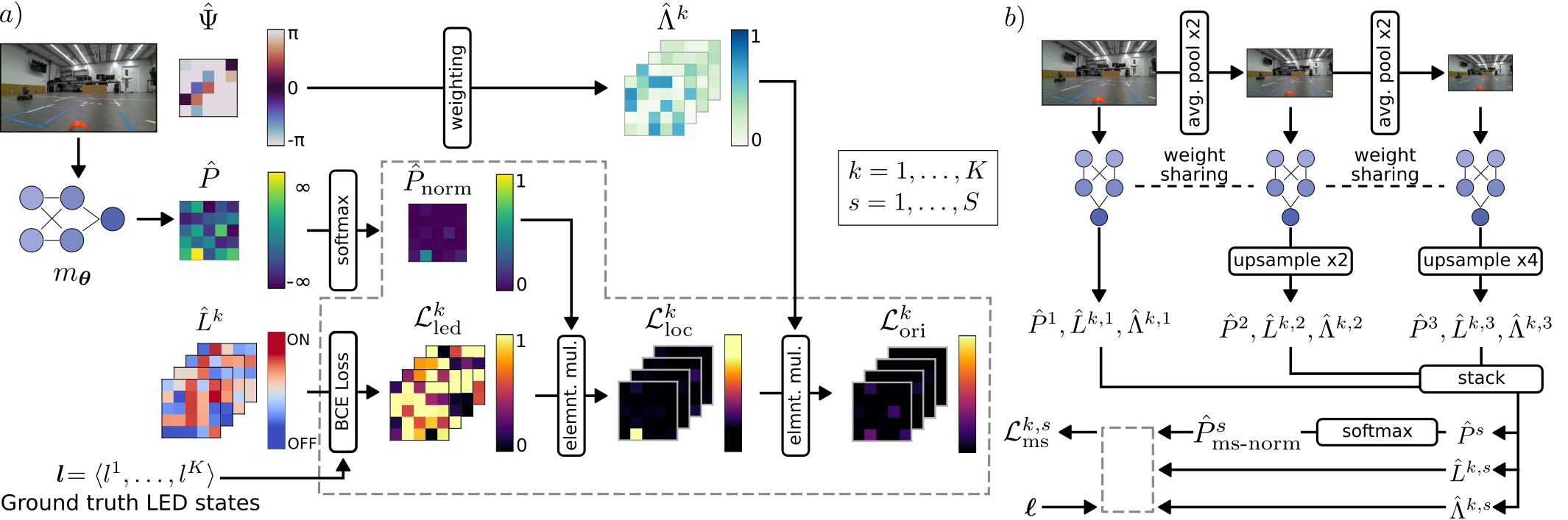}
    \caption{Overview of the approach: (a) given an input image, our approach predicts the robot's location in the image and its bearing relative to the camera. (b) We apply this mechanism over multiple rescaled versions of the input image to infer the robot's distance to  the camera.}
    \label{fig:method}
\end{figure}

\paragraph{Robot Localization.}
Most cells in $\hat{L}^k$ cannot predict the correct LED state because their limited \ac{rf} does not capture the robot, leading to high values in the $\mathcal{L}_\text{led}^k$ loss maps.
An intuitive way to lower the loss is to give less weight to errors corresponding to areas not depicting a robot and give more weight to the cells that see the robot (i.e., have the robot inside the \ac{rf}) as they can predict the LED states.
Thus, we allow the model to spatially weight the $\mathcal{L}_\text{led}^k$ maps through the $\hat{P}$ map;
each cell in $\hat{P}$ takes values in the $[0, 1]$ range, and indicates the belief about the robot's presence inside its RF.
We normalize this map with the softmax function, denoted as $\hat{P}_\text{norm}$, and define the localization loss $\mathcal{L}_\text{loc}^k = \mathcal{L}_\text{led}^k \odot \hat{P}_\text{norm}$, where $\odot$ indicates the element-wise product;
the softmax prevents the model from trivially setting the loss to zero.
With this formulation, the model is driven to produce high values in the cells of $\hat{P}$ whose \ac{rf} contains the robot, as these will generally correspond to the low loss cells of $\mathcal{L}_\text{led}^k$.
This weighting mechanism can be seen as spatial attention in CNNs~\citep{guo2022attention}, with the difference being that we apply it directly to a loss function instead of raw model features.

\begin{wrapfigure}{R}{.45\textwidth}
    \centering
    \includegraphics[width=0.9\linewidth]{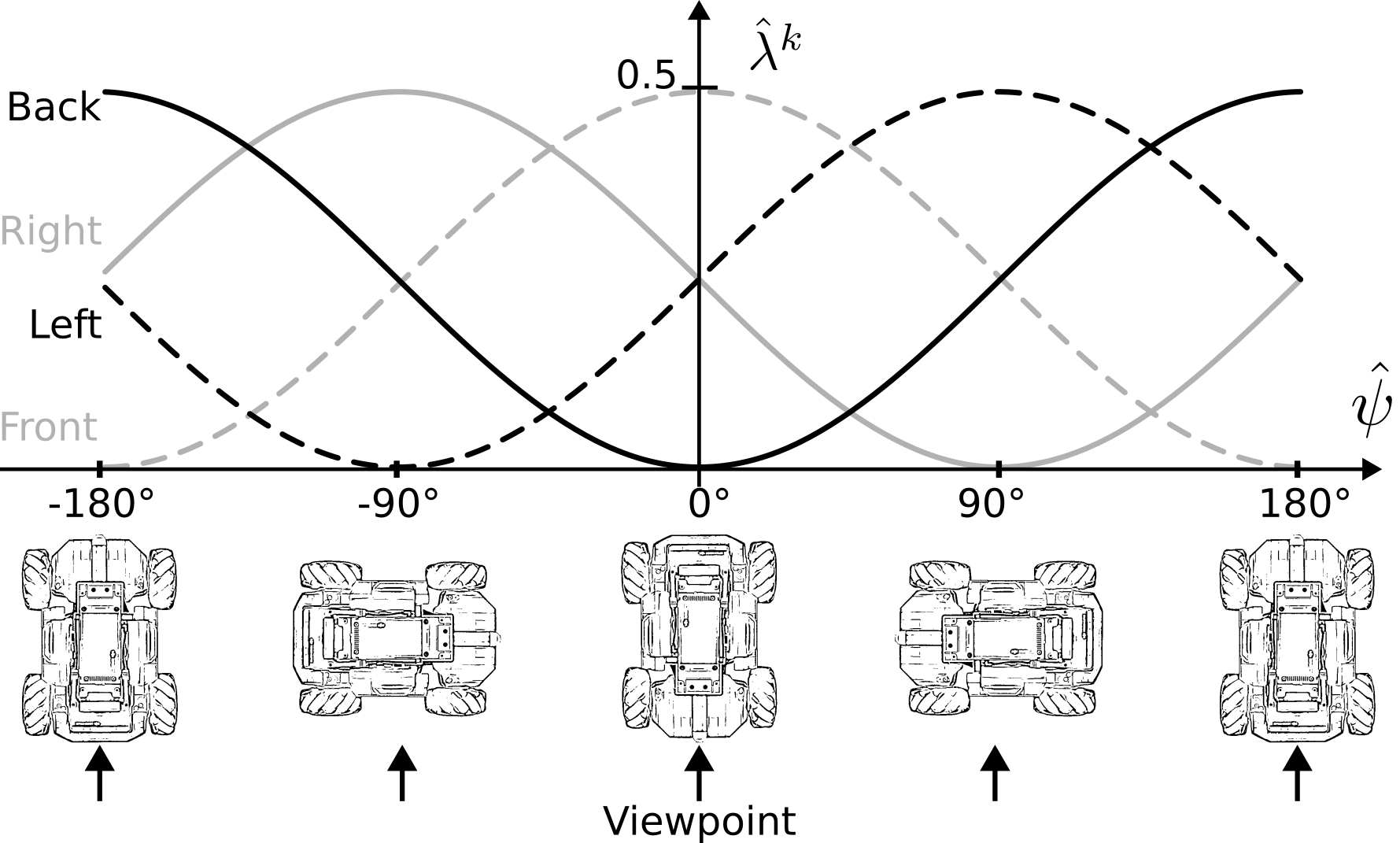}
    \caption{Visibility function for a robot with four LEDs at the cardinal directions. $\hat{\psi}$ is a cell of the predicted bearing map $\hat{\Psi}$, and $\hat{\lambda}^k$ is the visibility weight for each of the $K$ LEDs.}
    \label{fig:lambda}
\end{wrapfigure}

\paragraph{Robot Relative Bearing Estimation.}
Whenever the robot is visible in the image, some of its LEDs are occluded by its own body.
Since these LEDs are not visible, the model is unable to predict their state, contributing to high values in the $\mathcal{L}_\text{loc}^k$ maps.
We introduce a weighing mechanism for the localization loss based on the predicted robot's bearing; it allows the model to downplay errors caused by occluded LEDs as long as it correctly predicts the robot's bearing.
To accomplish this, we introduce the predicted robot's bearing map $\hat{\Psi}$, whose cells have values in the $[-\pi,\pi]$ range.
Each cell represents the robot's bearing relative to the camera, i.e., which side of the robot is visible, encoded as an angle (see Figure~\ref{fig:intro}).
We use a differentiable function to map the predicted bearing $\hat{\Psi}$ to $K$ visibility scores, one for each LED, resulting in the $\hat{\Lambda}^k$ maps.
Each element of $\hat{\Lambda}^k$ is defined as $\hat{\lambda}^k = \text{cos}( \hat{\psi} + \frac{2\pi}{K} (k - 1) )$, where $\hat{\psi}$ is an element of $\hat{\Psi}$; we designed $\hat{\Lambda}^k$ for $K$ equidistant LEDs mounted on the robot.
This visibility function reasonably approximates each LED's visibility from different viewing directions, though it is not precise nor the result of calibration.
We normalize the map values such that, given an orientation, all $K$ coefficients are non-negative and sum to one (see Figure~\ref{fig:lambda}).
The $\hat{\Lambda}^k$ maps are then used to compute the orientation loss $\mathcal{L}_\text{ori}^k = \mathcal{L}_\text{loc}^k \odot \hat{\Lambda}^k$.

\begin{figure}[t]
    \centering
    \includegraphics[width=\linewidth]{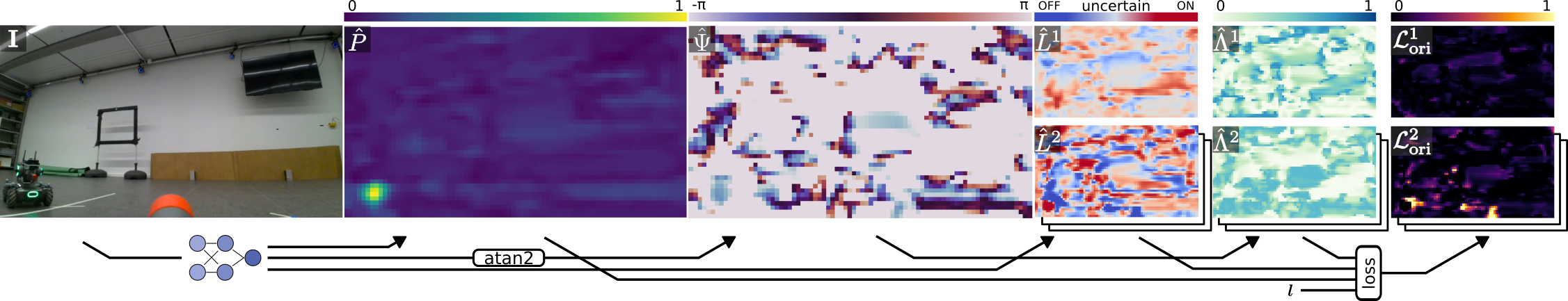}
    \caption{Model's output maps (a single scale was selected for visualization purposes). The $\hat{P}$ map represents the robot's presence, $\hat{\Psi}$ its bearing, and $\hat{L}^k$ the LED states (1 for back, 2 for front).}
    \label{fig:model-outputs}
\end{figure}

\paragraph{Robot Distance Estimation.}
Given an image, when the robot's apparent size and that of the RF do not match well, it interferes with correctly predicting the LED states: 
if the robot appears much larger than the \ac{rf}, the cell will miss contextual information about the robot's structure; if it is much smaller, the \ac{rf} will contain unrelated background information as the robot will be represented by few pixels.
We address this problem and exploit its solution to estimate the relative robot's distance, as the apparent image size of an object is directly proportional to its distance from the camera:
we pass the same image at different scales $\{ 
s_1, \dots, s_S \}$ to the model and consider the output maps $\hat{L}^{k,s}$, $\hat{P}^s$, $\hat{\Psi}^s$ from each forward pass, where $k$ is an LED and $s$ an image scale, as shown in Figure~\ref{fig:method}.
The output maps from the different passes are upscaled to have matching spatial dimensions.
Then, we compute the normalized multi-scale localization map $\hat{P}_\text{ms-norm} = \text{softmax}(\hat{P}^{s_1,\ldots,s_S})$.
We combine the previous loss with the multi-scale formulation and define the loss $\mathcal{L}_\text{ms}^{k,s} = \text{BCE}(\hat{L}^{k,s}, l^k) \odot \hat{P}_\text{ms-norm}^{s} \odot \hat{\Lambda}^{k,s}$.
%
%
By normalizing over all image scales, we enable the model to make a convex combination of scales, ensuring the robot size best fits into the combined multi-scale \ac{rf}s.
The coefficients of the combinations are used to compute the robot's distance from the camera.

Finally, we define the complete loss function in Equation~\eqref{eq:loss:final} to obtain a scalar loss value.
It computes the average over the spatial, LED, and scale dimensions of the multi-scale loss, where $X[i, j]$ is the indexing operator accessing the value of the cell at row $i$ and column $j$ of a generic map $X$.
Note that this loss accounts for all aspects captured by $\mathcal{L}_{ori}$, reformulated in multi-scale fashion.
\begin{equation}\label{eq:loss:final}
\mathcal{L} = \frac{1}{K} \sum_{k=1}^{K} \sum_{s=1}^{S} \sum_{i=1}^{H'} \sum_{j=1}^{W'} \mathcal{L}_\text{ms}^{k,s} [i, j]
\end{equation}
\paragraph{Model Inference.}
We compute the predicted robot location in the image $(\hat{u}, \hat{v})$ as the barycenter of the $\hat{P}_\text{ms-norm}^s$ maps.
The predicted relative bearing is the weighted average of $\hat{\Psi}^s$ by $\hat{P}_\text{ms-norm}^s$ and is directly mapped to the robot's orientation.
Lastly, we estimate the robot's distance by measuring its apparent size in the original input image, which is proportional to its physical distance from the camera.
We identify the robot's apparent size in the image by summing each $\hat{P}_\text{ms-norm}^s$ map over its spatial dimensions, obtaining a vector of $S$ elements that sum to one.
Using this vector to apply a linear combination of scale factors, we obtain a scalar representing the robot's size in the original input image.
Similarly to Depth Anything~\citep{yang2024depth,yang2024depthv2}, the predicted distance requires a calibration: we multiply it with the calibration factor $d_c$ derived from a single image with a known robot distance.
Note that we do not explicitly handle external occlusions to the LEDs. However, our experiments show the model is resilient to partial and total occlusions caused by the camera's FOV.
Following the inference procedure, we report an average running time of 6.5ms per image (153 Hz) on a NVIDIA GeForce RTX 4080.
More details on model inference can be found in the appendix.

\section{Experimental setup}\label{sec:setup}
We apply our proposed approach to the DJI Robomaster S1 robot, a ground rover equipped with a monocular RGB camera with a resolution of $640 \times 360$ pixels mounted on top of a pan and tilt turret.
The robot features six multi-color LEDs, of which we consider four for our experiments: the two on the turret and the front and back ones on the robot's base  (see Figure~\ref{fig:intro}); 
the left and right LEDs of the base are always turned off and ignored during training and evaluation.

\begin{figure}[t]
    \centering
    \includegraphics[width=\linewidth]{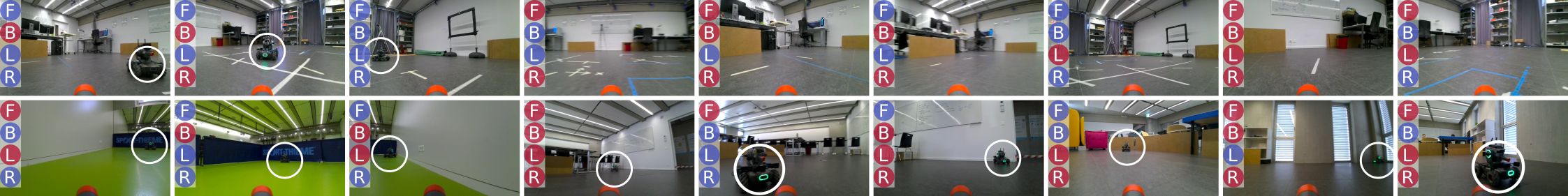}
    \caption{Random training samples from the datasets: laboratory (top row), gym (1-3 on bottom), classroom (4-6 on bottom), and break room (7-9 on bottom). On the right, the LED state ({\color{cool}blue} for off, {\color{warm} red} for on; F for front, B for back, L for left, R for right) is reported for each LED; white circles mark the robot. Only 23\% of collected samples feature a visible robot in the laboratory training set.}
    \label{fig:dataset}
\end{figure}

\paragraph{Data Collection.}
We let the two robots randomly move in different environments and independently randomize the state of each of the four LEDs every five seconds. Each robot broadcasts its LED states which are used as labels for the images acquired by the other one.
In total, the robots collected 131K samples in a laboratory environment, which are sequentially split into the the training $\mathcal{T}_\text{lab}$ (116K samples), validation $\mathcal{V}_\text{lab}$ (10K samples), and testing $\mathcal{Q}_\text{lab}$ (5K samples) sets.
Having no access to pose labels while randomly exploring, no measures are taken to ensure that the robots are in each other's \ac{fov};
consequently, 77\% of the training-set images depict an empty background with no robots, as shown in the top row of Figure~\ref{fig:dataset}.
To validate the model and assess its performance, the pose of both robots is collected using a motion capture system.
Our quantitative evaluation is carried out on the subset of testing set samples with a visible robot, amounting to 1K samples, denoted as $\mathcal{Q}_\text{lab}^\nu$.
We stress that pose information is not made available to any of our models.
However, we consider a supervised upperbound to measure the maximum performance achievable with our setup.

Additionally, data is collected in less unstructured environments, having no external tracking system and, as a consequence, no ground truth poses.
In detail, the robots collected 34K samples in a classroom, 48K samples in a gym, and 45k samples in a break room.
We combine samples from the three environments to create the \ac{ood} dataset, where we split data into a training set $\mathcal{T}_\text{ood}$ (120K samples), validation set $\mathcal{V}_\text{ood}$ (5K samples), and testing set $\mathcal{Q}_\text{ood}$ (2K samples).

\paragraph{Baselines.}
We compare our model against \textit{Mean Predictor},  \textit{CNOS}~\citep{nguyen2023cnos}, and \textit{Upperbound} models:
\textit{Mean Predictor} always returns the mean relative pose of the robot in the laboratory training set.
\textit{Upperbound} is a version of our architecture trained in a fully supervised fashion, i.e., it represents the assumption that pose labels are available for every image and trains with pose labels in the laboratory training set, denoted as $\mathcal{T}_\text{lab}^*$.
\textit{CNOS} is a SoA novel object pose estimation approach based on the CAD model of the object of interest;
it segments the input image using SAM~\citep{kirillov2023segment},  matches the segmentations to rendered templates using the features from DINOv2~\citep{oquab2023dinov2}, and returns the known pose of the matched template as its prediction.
Specifically, we generate 400 templates by rendering the robot's textured CAD model at 4 distance settings (0.5m, 1m, 2m, 4m) across 100 different orientations.
We also consider MegaPose\citep{labbe2023megapose} as a baseline; similarly to \textit{CNOS}, it matches the input image with templates of the robot's CAD model, and uses the recovered pose as an initial pose guess refined with a render \& compare strategy. However, its quantitative performance is worse than all other baselines considered and takes more than a minute to infer the pose from an image.

\paragraph{Evaluation Metrics.}
All metrics are computed on  $\mathcal{Q}_\text{lab}^\nu$:
for localization, we measure the median distance of the robot's center pixel location, called $E_{uv}$;
for the orientation, we compute the median circular error~\citep{mardia2009directional}, called $E_\psi$;
for the distance, we measure the mean absolute percentage error of the distance --which is not influenced by the distance distribution in the dataset compared to the absolute error--, called $E_d$, and defined as $\scaleto{\sfrac{| d - \hat{d}|}{d}}{12pt}$;
for the LED classification, we measure the AUC averaged over the LEDs that are visible according to the ground truth robot pose.
%
%
Similarly to SO-Pose~\citep{di2021sopose}, we measure the overall goodness of our approach with the pose accuracy metric represented as $\Gamma^{45^\circ}_{1\text{m}}$ and defined as the percentage of predictions with a position error of less than $1$ meters and orientation error of less than $45^\circ$ from the ground truth pose.
%
%
The threshold values are heuristically set such that \textit{Upperbound} has a score greater than $90\%$ in the $\Gamma^{45^\circ}_{1\text{m}}$ metric.

\section{Results}\label{sec:results}
We evaluate the pose estimation performance of our approach by training our model on the laboratory training dataset and evaluating it on the testing set collected in the same environment.
For evaluating the model, we ignore empty scenes in the testing set.
\BeforeBeginEnvironment{wrapfigure}{\setlength{\intextsep}{1mm}}%
\begin{wrapfigure}{R}{.43\textwidth}
    \centering
    \begin{tikzpicture}[background rectangle/.style={fill=red!20}]%
%
\def\scatterlen{32mm}
\begin{scope}
\clip(-1.1,-2.85) rectangle (4.15,2.1);
\begin{axis}[%
 name=scatter_u,
 axis equal,
 axis on top,
 height=\scatterlen,
 width=\scatterlen,
 title={$u$[px]},
 ylabel={Prediction},
 title style={yshift=-2mm,},
 xmin=0.0, xmax=1.0,
 ymin=0.0, ymax=1.0,
 xtick={0.00, 0.25, 0.50, 0.75, 1.00},
 ytick={0.00, 0.25, 0.50, 0.75, 1.00},
 minor x tick num=1,
 minor y tick num=1,
 xticklabels=\empty,
 yticklabels={0, 160, 320, 480, 640},
 xtick pos=left,
 ytick pos=bottom,
 ticklabel style={font=\footnotesize},
 xlabel style={font=\small},
 ylabel style={font=\small},
]%
\addplot graphics[xmin=\pgfkeysvalueof{/pgfplots/xmin},ymin=\pgfkeysvalueof{/pgfplots/ymin},xmax=\pgfkeysvalueof{/pgfplots/xmax},ymax=\pgfkeysvalueof{/pgfplots/ymax}] {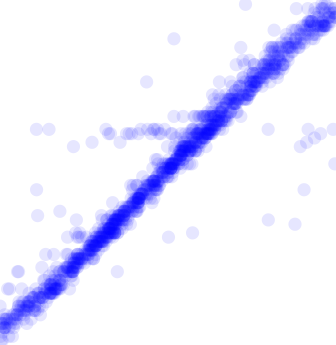};
\addplot[color=wong_black,line width=1.0pt,opacity=0.4,dotted] coordinates {(\pgfkeysvalueof{/pgfplots/xmin},\pgfkeysvalueof{/pgfplots/ymin}) (\pgfkeysvalueof{/pgfplots/xmax},\pgfkeysvalueof{/pgfplots/ymax})};%
\end{axis}%
\begin{axis}[%
 name=scatter_v,
 axis equal,
 axis on top,
 at=(scatter_u.right of north east),
 anchor=left of north west,
 height=\scatterlen,
 width=\scatterlen,
 title={$v$[px]},
 title style={yshift=-2mm,},
 xmin=0.0, xmax=1.0,
 ymin=0.0, ymax=1.0,
 xtick={0.00, 0.25, 0.50, 0.75, 1.00},
 ytick={0.00, 0.25, 0.50, 0.75, 1.00},
 minor x tick num=1,
 minor y tick num=1,
 xticklabels=\empty,
 yticklabels={0, 90, 180, 270, 360},
 xtick pos=left,
 ytick pos=bottom,
 ticklabel style={font=\footnotesize},
 xlabel style={font=\small},
 ylabel style={font=\small},
]%
\addplot graphics[xmin=\pgfkeysvalueof{/pgfplots/xmin},ymin=\pgfkeysvalueof{/pgfplots/ymin},xmax=\pgfkeysvalueof{/pgfplots/xmax},ymax=\pgfkeysvalueof{/pgfplots/ymax}] {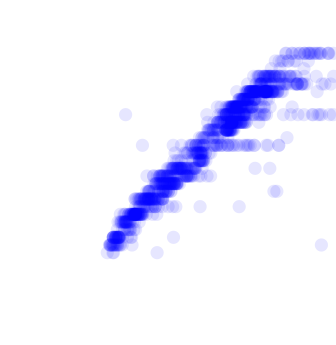};
\addplot[color=wong_black,line width=1.0pt,opacity=0.4,dotted] coordinates {(\pgfkeysvalueof{/pgfplots/xmin},\pgfkeysvalueof{/pgfplots/ymin}) (\pgfkeysvalueof{/pgfplots/xmax},\pgfkeysvalueof{/pgfplots/ymax})};%
\end{axis}%
\begin{axis}[%
 name=scatter_psi,
 axis equal,
 axis on top,
 at=(scatter_u.below south west),
 anchor=above north west,
 height=\scatterlen,
 width=\scatterlen,
 title={$\psi$[deg]},
 title style={yshift=-2mm,},
 xlabel={Ground Truth},
 ylabel={Prediction},
 ylabel shift=-1mm,
 xmin=0.0, xmax=1.0,
 ymin=0.0, ymax=1.0,
 xtick={0.00, 0.25, 0.50, 0.75, 1.00},
 ytick={0.00, 0.25, 0.50, 0.75, 1.00},
 minor x tick num=1,
 minor y tick num=1,
 xticklabels=\empty,
 yticklabels={-180,-90,0,90,180},
 xtick pos=left,
 ytick pos=bottom,
 ticklabel style={font=\footnotesize},
 xlabel style={font=\small, yshift=2mm,},
 ylabel style={font=\small},
]%
\addplot graphics[xmin=\pgfkeysvalueof{/pgfplots/xmin},ymin=\pgfkeysvalueof{/pgfplots/ymin},xmax=\pgfkeysvalueof{/pgfplots/xmax},ymax=\pgfkeysvalueof{/pgfplots/ymax}] {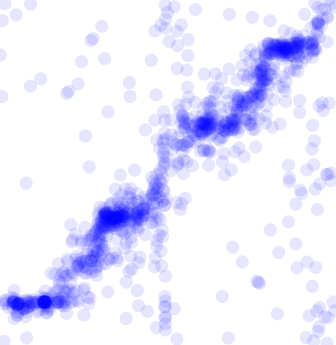};
\addplot[color=wong_black,line width=1.0pt,opacity=0.4,dotted] coordinates {(\pgfkeysvalueof{/pgfplots/xmin},\pgfkeysvalueof{/pgfplots/ymin}) (\pgfkeysvalueof{/pgfplots/xmax},\pgfkeysvalueof{/pgfplots/ymax})};%
\end{axis}%
\begin{axis}[%
 name=scatter_dist,
 axis equal,
 axis on top,
 at=(scatter_v.below south west),
 anchor=above north west,
 height=\scatterlen,
 width=\scatterlen,
 title={$d$[m]},
 title style={yshift=-2mm,},
 xlabel={Ground Truth},
 xmin=0.0, xmax=1.0,
 ymin=0.0, ymax=1.0,
 xtick={0.00, 0.25, 0.50, 0.75, 1.00},
 ytick={0.00, 0.25, 0.50, 0.75, 1.00},
 minor x tick num=1,
 minor y tick num=1,
 xticklabels=\empty,
 yticklabels={0, 1, 2, 3, 4},
 xtick pos=left,
 ytick pos=bottom,
 ticklabel style={font=\footnotesize},
 xlabel style={font=\small, yshift=2mm,},
 ylabel style={font=\small},
]%
\addplot graphics[xmin=\pgfkeysvalueof{/pgfplots/xmin},ymin=\pgfkeysvalueof{/pgfplots/ymin},xmax=\pgfkeysvalueof{/pgfplots/xmax},ymax=\pgfkeysvalueof{/pgfplots/ymax}] {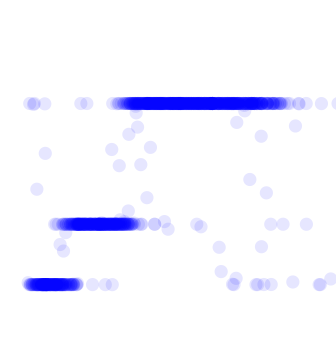};
\addplot[color=wong_black,line width=1.0pt,opacity=0.4,dotted] coordinates {(\pgfkeysvalueof{/pgfplots/xmin},\pgfkeysvalueof{/pgfplots/ymin}) (\pgfkeysvalueof{/pgfplots/xmax},\pgfkeysvalueof{/pgfplots/ymax})};%
\end{axis}%
\end{scope}
\end{tikzpicture}%
    \caption{\textit{Our} self-supervised model predictions vs ground truth on $\mathcal{Q}_\text{lab}^\nu$. Our approach discretizes distances into three bins, resulting in a coarse step function. 
        }
    \label{fig:detection-matrices}
\end{wrapfigure}
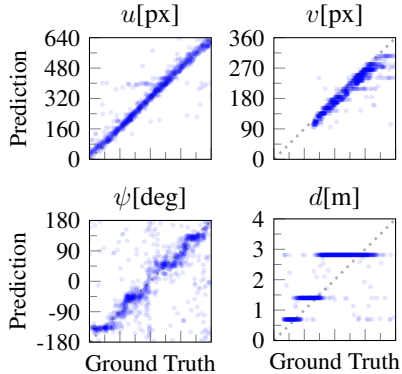
We report the results in Table~\ref{tab:table}, where we observe that our self-supervised model significantly better than the baselines despite requiring to be trained only on a dataset of images labeled with the binary state of four LEDs, needing no auxiliary segmentation model or rendered CAD templates.

Our model's performance largely surpasses the \textit{Mean Predictor} and closely follows the \textit{Upperbound}.
The performance gap with the latter is in the distance estimation; this is explained by the fully-supervised approach regressing the robot's distance as a continuous value, as opposed to our method relying on a discrete number of scales for the prediction, as depicted in Figure~\ref{fig:detection-matrices}.
In preliminary experiments, we verified that the issue can be mitigated by using more scales during inference.
The \textit{CNOS} approach has the lowest $E_{uv}$ error, thanks to the performance of SAM~\citep{kirillov2023segment} and access to the robot's CAD model. However, it struggles the most in predicting the distance and heading. 
Upon inspecting the problem, we discovered that DINOv2's embeddings~\citep{oquab2023dinov2} have similar values for a rendered robot template and its horizontal reflection, and for the same pose at different distances.

\begin{table*}[t]
    \setlength\tabcolsep{1.2mm} 
    \renewcommand{\arraystretch}{1.0} 
    \centering
    \caption{Performance metrics computed on the laboratory testing set $\mathcal{Q}_\text{\textup{lab}}^\nu$, three replicas per row.}
    \begin{tabular}{lcrrrrrr>{\centering\arraybackslash}p{40mm}}
    \toprule
    \multirow{2}{*}{Model} & \multirow{2}{*}{Supervision} & \multicolumn{1}{c}{$E_{uv}$} & \multicolumn{1}{c}{$E_{\psi}$} & \multicolumn{1}{c}{$E_{d}$} & \multicolumn{1}{c}{$\Gamma^{\ang{45}}_{1\text{m}}$} & \multicolumn{1}{c}{AUC} & {Point plot for $\Gamma^{\ang{45}}_{1\text{m}}$ [\%] $\rightarrow$} \\
    & & [px] $\downarrow$ & [deg] $\downarrow$ & [\%] $\downarrow$ & \multicolumn{1}{c}{[\%] $\uparrow$} & [\%] $\uparrow$  & {Error bars mark 95\% CI} \\
    \midrule
    \textit{Mean Predictor}              & $\mathcal{T}_\text{lab}$   & 141 & 86 & 34 & 10 &  50 & \multirow{1}{*}{\input{tableplot.tikz}} \\
    \textit{Ours}      & $\mathcal{T}_\text{lab}$   &  17 & 17 & 24 & 70 &  98 & \\
    \textit{Ours} (OOD)  & $\mathcal{T}_\text{ood}$   &  27 & 55 & 60 & 29 &  89 & \\
    \textit{CNOS}~\citep{nguyen2023cnos} & CAD model             &  13 & 72 & 35 & 25 & N/A & \\
    \textit{Upperbound}                  & $\mathcal{T}^*_\text{lab}$ &  18 & 14 & 11 & 93 &  99 & \\[1mm]
    \bottomrule
    \end{tabular}
    \label{tab:table}
\end{table*}

\paragraph{LEDs are not necessary at deployment time}\label{sec:results:table:led}
We test our model on the subset of $\mathcal{Q}_\text{lab}^\nu$ with images having all visible LEDs turned off, where it scores $E_{uv}~=~19.6\text{px}$, $E_{\psi}=\ang{19.23}$, and $E_{d} = 25.7\text{\%}$.
The small difference in performance compared to testing on the entire $\mathcal{Q}_\text{lab}^\nu$ demonstrates that the approach is robust and does not require a specific LED state for accurate pose estimation. Higher errors are attributed to the reduced visibility of the robot when LEDs are off.
\paragraph{Fine Tuning to Novel Environments.}\label{sec:results:finetuning}
We consider a model pre-trained using our approach on $\mathcal{T}_\text{ood}$ and fine-tuned (using our approach) on an increasing number of samples from the laboratory (from 5K to 115K).
%
%
The third row of Table~\ref{tab:table} reports the performance of the model pre-trained on OOD data and later tested on $\mathcal{Q}_\text{lab}^{\nu}$ without fine-tuning;
the performance of the model is better than the mean predictor, while it struggles with distance estimation.
\BeforeBeginEnvironment{wrapfigure}{\setlength{\intextsep}{-1mm}}%
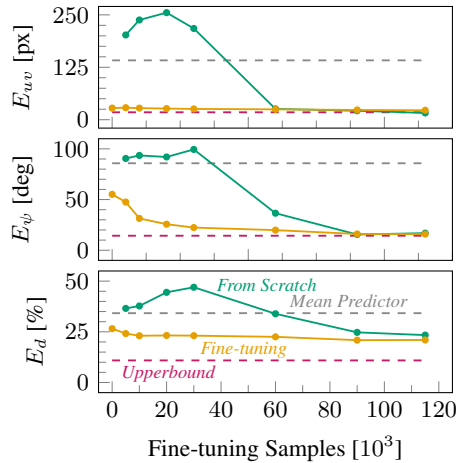
\begin{wrapfigure}{r}{.45\textwidth}
    \centering
    \def\finetuneheight{32mm}
\begin{tikzpicture}%
\begin{axis}[%
 name=tuning_uv,
 height=\finetuneheight,
 width=\linewidth,
 ylabel={$E_{uv}$ [px]},
 xmin=-5, xmax=125,
 ymin=-20, ymax=270,
 ytick={0, 125, 250},
 minor x tick num=3,
 minor y tick num=4,
 xticklabels=\empty,
 xtick pos=left,
 ytick pos=bottom,
 ticklabel style={font=\footnotesize},
 xlabel style={font=\small},
 ylabel style={font=\small},
]%
\addplot[color=wong_gray,line width=0.7pt,dashed] table[x=samples,y=mean_uv] {tuning.csv};%
\addplot[color=wong_magenta,line width=0.7pt,dashed] table[x=samples,y=upper_uv] {tuning.csv};%
\addplot[color=wong_green,line width=0.7pt,mark=*,mark size=1pt,restrict x to domain=1:120] table[x=samples,y=scratch_uv] {tuning.csv};%
\addplot[color=wong_gold,line width=0.7pt,mark=*,mark size=1pt] table[x=samples,y=tuning_uv] {tuning.csv};%
\end{axis}%
\begin{axis}[%
 name=tuning_psi,
 at=(tuning_uv.below south),
 yshift=1mm,
 anchor=north,
 height=\finetuneheight,
 width=\linewidth,
 ylabel={$E_{\psi}$ [deg]},
 xmin=-5, xmax=125,
 ymin=-10, ymax=110,
 ytick={0, 50, 100},
 minor x tick num=3,
 minor y tick num=4,
 xticklabels=\empty,
 xtick pos=left,
 ytick pos=bottom,
 ticklabel style={font=\footnotesize},
 xlabel style={font=\small},
 ylabel style={font=\small},
]%
\addplot[color=wong_gray,line width=0.7pt,dashed] table[x=samples,y=mean_psi] {tuning.csv};%
\addplot[color=wong_magenta,line width=0.7pt,dashed] table[x=samples,y=upper_psi] {tuning.csv};%
\addplot[color=wong_green,line width=0.7pt,mark=*,mark size=1pt,restrict x to domain=1:120] table[x=samples,y=scratch_psi] {tuning.csv};%
\addplot[color=wong_gold,line width=0.7pt,mark=*,mark size=1pt] table[x=samples,y=tuning_psi] {tuning.csv};%
\end{axis}%
\begin{axis}[%
 name=tuning_dist,
 at=(tuning_psi.below south),
 yshift=1mm,
 anchor=north,
 height=\finetuneheight,
 width=\linewidth,
 xlabel={Fine-tuning Samples [$10^3$]},
 ylabel={$E_{d}$ [\%]},
 xmin=-5, xmax=125,
 ymin=-5, ymax=55,
 xtick={0, 20, 40, 60, 80, 100, 120},
 ytick={0, 25, 50},
 minor x tick num=1,
 minor y tick num=4,
 xtick pos=left,
 ytick pos=bottom,
 ticklabel style={font=\footnotesize},
 xlabel style={font=\small},
 ylabel style={font=\small},
]%
\addplot[color=wong_gray,line width=0.7pt,dashed] table[x=samples,y=mean_dist] {tuning.csv};%
\addplot[color=wong_magenta,line width=0.7pt,dashed] table[x=samples,y=upper_dist] {tuning.csv};%
\addplot[color=wong_green,line width=0.7pt,mark=*,mark size=1pt,restrict x to domain=1:120] table[x=samples,y=scratch_dist] {tuning.csv};%
\addplot[color=wong_gold,line width=0.7pt,mark=*,mark size=1pt] table[x=samples,y=tuning_dist] {tuning.csv};%
\node at (axis cs:62,40) [font={\scriptsize},anchor=mid west] {\color{wong_gray}\textit{Mean Predictor}};%
\node at (axis cs:0,5) [font={\scriptsize},anchor=mid west] {\color{wong_magenta}\textit{Upperbound}};%
\node at (axis cs:35,48) [font={\scriptsize},anchor=mid west] {\color{wong_green}\textit{From Scratch}};%
\node at (axis cs:29,17) [font={\scriptsize},anchor=mid west] {\color{wong_gold}\textit{Fine-tuning}};%
\end{axis}%
\end{tikzpicture}%
    \caption{Metrics of a model trained from scratch on $\mathcal{T}_\text{lab}$ (in {\color{wong_green}green}) and a model pre-trained on $\mathcal{T}_\text{ood}$ and fine-tuned on $\mathcal{T}_\text{lab}$ (in {\color{wong_gold!90!black}yellow}). The performance of \textit{Mean Predictor} (in {\color{wong_gray!80!black}gray}) and \textit{Upperbound} (in {\color{wong_magenta}magenta}) are reported as dashed lines for comparison.}
    \label{fig:fine-tuning}
\end{wrapfigure}
We further fine-tune the model on the target domain, represented by the \textit{Fine-tuning} model, and compare with another trained from scratch on the same data, called \textit{From Scratch}, in Figure~\ref{fig:fine-tuning}.
We observe that training from scratch with less than 30K fine-tuning samples achieves a worse performance than using a pre-trained model in different environments, highlighting
the model's generalization ability.
We note that the amount of data needed for the from-scratch model to achieve similar performance to the fine-tuned one depends on the task: localizing the robot is easier than estimating its heading which, in turn, is easier than estimating the robot's distance.

\begin{figure}[t]
    \centering
    \includegraphics[width=1\linewidth]{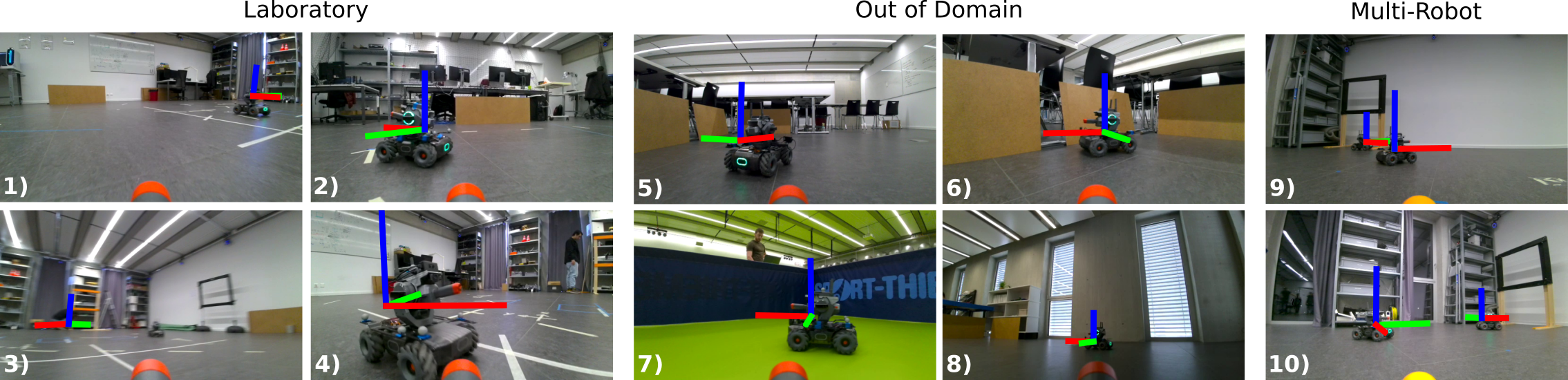}
    \caption{Predicted robot poses (x-axis in {\color{red}red}, y-axis in {\color{green!60!black}green}, z-axis in {\color{blue}blue}) by models trained with our approach: a model trained on $\mathcal{T}_\text{lab}$ applied to $\mathcal{Q}_\text{lab}^\nu$ (1-4); a model trained on $\mathcal{T}_\text{ood}$ and applied to the testing set in classroom (5, 6), gym (7), and break room (8); a model trained on $\mathcal{T}_\text{lab}$ applied to images with multiple robots (9, 10). Large errors occur when the images are blurred or robots are far from the camera (3), while smaller errors occur when robot and background blend together (4).}
    \label{fig:res.qualitative}
\end{figure}

\paragraph{The Model is Capable of Multi-robot Pose Estimation.}
We show model predictions on unseen images featuring multiple robots in Figure~\ref{fig:res.qualitative}. Despite being trained on images with at most one visible robot, the approach correctly works with images with multiple ones. In this experiment, each robot corresponds to a different local maximum of $\hat{P}$.
Details on adapting the model inference for multi-robot pose estimation can be found in the appendix.

\setlength{\intextsep}{5mm}%

\section{Conclusions}\label{sec:conclusions}
We presented a self-supervised pose estimation approach that does not require pose labels; instead, supervision is obtained from classifying the state of multiple, independent LEDs on the robot's body.
A pair of robots collect images and the ground truth LED state of the peer autonomously, without external hardware, lending the approach to online learning and domain adaptation.
Results indicate that our approach trains a competitive pose predictor, whose performance is not degraded by the lack of pose labels despite the complexity of the task.

Our approach only assumes that changes in the target variable (e.g., state of the LEDs) affect the appearance of the robot such that the sensor (e.g., monocular camera) can observe and predict this change of state.
As such, the approach can be applied to different sensors, and other actuators (e.g. the position of an arm), as long as it affects the robot's appearance in the sensed data.

\section{Limitations}

Our model and experiments focus on 2D pose estimation for ground robots.
Extension to 3D would be trivial for the position component of the pose; for the rotation component, the LED visibility function would be defined on the space of unit quaternions (representing 3D rotations).
In order to disambiguate robot pitch, additional LEDs would be required that face in directions with positive and negative pitches.
Reconstructing the roll would require considering the detected locations of each specific LED, which are accessible in the raw LED maps.

The laboratory testing set used in our quantitative experiments, despite being cluttered and full of distractors on shelves, does not contain images where the target robot is occluded by other objects; as such, the efficacy of the approach on partial occlusions needs further testing.

Our current formulation relies on rescaling an input image multiple times (three in our experiments), running a forward pass on each scaled version, and combining the predictions to get the estimated distance.
The accuracy of the resulting distance estimate is limited by the coarse discretization of the image scales used to estimate the distance. At inference time, this can be mitigated by considering more scaling factors at the expense of additional computation.
In future work, we plan to explore different multi-scale architectures, such as U-Net models~\citep{ronneberger2015u}, to address this limitation and improve the pose estimation performance.  
Further, our approach considers each training frame individually, disregarding valuable temporal information such as the robot's odometry or the image optical flow.
The approach can benefit from incorporating this information in the training process, e.g., by employing an auxiliary consistency loss~\citep{nava2021state} between pairs of frames.

The training approach assumes to have at most one robot inside the \ac{fov}, limiting its application to a pair of robots collecting data.
We plan to extend the approach to handle large groups of collaborative robots, dramatically improving data collection efficiency and increasing the frames featuring visible robots (23\% in our training set): the amount of useful collected data over a given time interval scales quadratically with the number of robots simultaneously deployed in the environment. 

The main limitation of our approach compared to CAD-based ones is the need to retrain from scratch for every new robot whose pose is to be estimated.
This drawback is mitigated by the ease of collecting data and the advantage of training directly on real-world images, eliminating the sim-to-real gap.


\clearpage

\bibliography{references}

\newpage
\renewcommand*{\thesection}{\Alph{section}}
\setcounter{section}{0}
\section*{Appendix}

\section{Neural Network Training}
Throughout this document, we adopt the same \ac{fcn} architecture, receiving as input an image with an original resolution of $640 \times 360$ pixels and producing maps of $80 \times 45$ cells, with a \ac{rf} of $70 \times 70$ pixels.
In detail, we designed a lightweight \ac{fcn} with 6 blocks interleaved by 2x max-pooling and totaling 179K parameters; each block consists of a 2D convolution, batch normalization, and ReLU non-linearity.
Our approach uses the scaling factors $(1, \frac{1}{2}, \frac{1}{4})$ to rescale the original input image with average pooling and, for each one, does a forward pass to produce the output maps.
After the three forward passes, we upscale the smaller maps to match the $80 \times 45$ size of the largest map (which corresponds to the largest scale) using bilinear interpolation.
Given these maps, we optimize the loss function defined in Equation~\eqref{eq:loss:final} using Adam~\citep{adam} with an initial learning rate $\eta_\text{initial} = 1e^{-3}$ that smoothly decreases to $\eta_\text{final} = 1e^{-4}$ with cosine interpolation~\citep{loshchilov2017sgdr} over 100 training epochs.
During training, we apply image augmentation using multiplicative simplex noise and color jittering.
The best set of parameters $\bm{\theta}$ for the model is chosen as the one leading to the smallest validation loss, usually occurring within the first 60 epochs of training.

\section{Model Inference}
The target robot location in the image $\hat{u}$, $\hat{v}$ is computed from $\hat{P}_\text{ms-norm}$:
at first, we sum the map cells over the scales $S$ to have a two-dimensional map whose cells indicate the presence of robots across all scales;
secondly, we compute the barycenter of the map by multiplying the values of each cell by its integer coordinates and summing all cells to obtain $\hat{u}'$, $\hat{v}'$;
finally, we localize the robot by multiplying $\hat{u}'$, $\hat{v}'$ by the integer factor relating the output maps' resolution to the original input image resolution.
Formally, the procedure can be written as
\begin{equation}\label{eq:inference:loc}
\begin{pmatrix} \hat{u} \\ \hat{v} \end{pmatrix} = 
\begin{pmatrix} \sfrac{W}{W'} \\ \sfrac{H}{H'} \end{pmatrix}
\sum_{s=1}^{S} \sum_{i=1}^{H'} \sum_{j=1}^{W'} ( \hat{P}_\text{ms-norm}^{s} \odot M )[i, j] 
\end{equation}
where $\odot$ is the element-wise product, and $M$ is the coordinate matrix consisting of cells $m_{ij} = \begin{pmatrix} i \  j \end{pmatrix}^T$.

The target robot orientation $\hat{\psi}$ is computed as the weighted average of $\hat{\Psi}^s$ by $\hat{P}_\text{ms-norm}^{s}$, defined as
\begin{equation}\label{eq:inference:ori}
\hat{\psi} = \sum_{s=1}^{S} \sum_{i=1}^{H'} \sum_{j=1}^{W'} ( \hat{P}_\text{ms-norm}^{s} \odot \hat{\Psi}^s )[i, j]
\end{equation}

Finally, the target robot distance is computed from $\hat{P}_\text{ms-norm}^{s}$ as follows:
\begin{equation}\label{eq:inference:dist}
\hat{d} = d_c\cdot\hat{d}'\,; \quad \quad \hat{d}' = \sum_{s=1}^{S} f_s \cdot \sum_{i=1}^{H'} \sum_{j=1}^{W'} ( \hat{P}_\text{ms-norm}^{s})[i, j]
\end{equation}
In this formula, we first sum the map cells over their width and height to have a one-dimensional vector whose elements indicate the degree of compatibility between the robot's image size and the model's \ac{rf} at different scales -- the higher the value, the better the robot size matches the \ac{rf}.
Recalling that the size of the robot's bounding box in the image is inversely proportional to its distance, we recover the distance $\hat{d}'$ as the average of the inverse of scale factors weighted by the vector defined above.
To get a metric prediction from $\hat{d}'$, we employ a simple calibration procedure: 
the robot is placed in front of the camera, and images of it are taken while adjusting its distance; we pick $d_c$ as the distance at which the robot appears with a size of $r \times r$ pixels (i.e., our model's \ac{rf}, as introduced in Section~\ref{sec:method}) in the captured images.

The target robot LED states $\hat{l}^k$ are computed as the average of $\hat{L}^{ks}$ weighted by $\hat{P}_\text{ms-norm}^{s}$, defined as
\begin{equation}\label{eq:inference:led}
\hat{l}^k = \sum_{s=1}^{S} \sum_{i=1}^{H'} \sum_{j=1}^{W'} ( \hat{P}_\text{ms-norm}^{s}  \circ \hat{L}^{ks} )[i, j]
\end{equation}

\section{Multi-robot Inference}
To allow the model to predict the pose of multiple robots, we modified the way the $\hat{P}$ map is obtained.
As described in Section~\ref{sec:method}, the cells in $\hat{P}$ are bound to the $[0, 1]$ value range.
This is achieved by applying a softmax function to the raw activation values of the map.
When multiple robots are present in the input image, the raw map contains multiple peak values at different spatial locations.
Because the softmax function suppresses all non-maxima peaks when the absolute difference between peaks is high enough, the post-softmax map generally presents only one peak.
Hence, our solution is first to linearly rescale the values into the $[0,1]$ and then to apply the softmax function to further suppress noise.

\section{Detecting Robot Presence}\label{sec:results:table:detection}

Taking inspiration from Weakly Supervised Learning, we explore how to use our learned model for robot detection.
We consider the problem of classifying whether a robot is visible anywhere in an image; we take the maximum of $\hat{P}_\text{ms-norm}$ as the belief about the presence of a robot.  When testing this binary classification approach over $\mathcal{Q}_\text{lab}$, we obtain an AUC of $83.4\%$.
Additionally, we consider a method that estimates robot presence based on the model's confidence in its LED predictions based the entropy formula:
\begin{equation}
    \frac{1}{4}\sum_{k=1}^41-\Bigl(-\hat{l}^k\cdot \operatorname{log}(\hat{l}^k) - (1 - \hat{l}^k)\cdot \operatorname{log}(1 - \hat{l}^k)\Bigl)
\end{equation}
The resulting scalar measures the model's confidence in its LED state prediction; the closer the values are to zero (off) or one (on), the higher the confidence.
Assuming that the model predicts the LED state with high confidence only when the robot is visible in the image, we use this value to detect the robot's presence.
Using this alternative method, we report a robot detection AUC of $97.2\%$.
We believe the difference between the two methods to be caused by the softmax operator applied to $\hat{P}_\text{ms-norm}$:
when no robot is visible, the softmax drastically accentuates the noise in the localization map, leading to many false positives. 
In this situation, our model produces LED maps with values close to 0.5, resulting in low confidence and, thus, stronger robot detection.

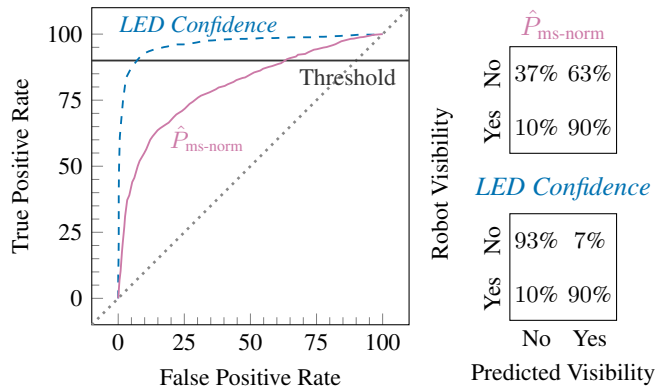
\begin{figure}[bh]
    \centering
    \begin{tikzpicture}%
\pgfplotsset{%
  colormap={allwhite}{rgb(0)=(1.0,1.0,1.0), rgb(255)=(1.0,1.0,1.0)},
}%
\def\confmatlen{30mm}%
\begin{axis}[%
 name=roc,
 axis equal,
 height=58mm,
 width=58mm,
 xlabel={False Positive Rate},
 ylabel={True Positive Rate},
 xmin=-0.1, xmax=1.1,
 ymin=-0.1, ymax=1.1,
 xtick={0.00, 0.25, 0.50, 0.75, 1.00},
 ytick={0.00, 0.25, 0.50, 0.75, 1.00},
 minor x tick num=4,
 minor y tick num=4,
 xticklabels={0, 25, 50, 75, 100},
 yticklabels={0, 25, 50, 75, 100},
 xtick pos=left,
 ytick pos=bottom,
 ticklabel style={font=\footnotesize},
 xlabel style={font=\small},
 ylabel style={font=\small},
]%
\addplot[color=wong_gray,line width=1.0pt,dotted] coordinates {(\pgfkeysvalueof{/pgfplots/xmin},\pgfkeysvalueof{/pgfplots/ymin}) (\pgfkeysvalueof{/pgfplots/xmax},\pgfkeysvalueof{/pgfplots/ymax})};%
\addplot[color=wong_black,line width=0.7pt] coordinates {(\pgfkeysvalueof{/pgfplots/xmin},0.90) (\pgfkeysvalueof{/pgfplots/xmax},0.90)};%
\addplot[color=wong_blue,line width=0.7pt,dashed] table[x=led_fpr,y=led_tpr] {detection.csv};%
\addplot[color=wong_pink,line width=0.7pt] table[x=pos_fpr,y=pos_tpr] {detection.csv};%
\node at (axis cs:0.16,0.59) [font={\footnotesize},anchor=mid west,align=center] {\color{wong_pink}$\hat{P}_\text{ms-norm}$};%
\node at (axis cs:-0.03,1.03) [font={\footnotesize},anchor=mid west,align=center] {\color{wong_blue}\textit{LED Confidence}};%
\node at (axis cs:0.65,0.83) [font={\footnotesize},anchor=mid west,align=center] {\color{wong_black}Threshold};%
\end{axis}%
\begin{axis}[%
 name=cm_entropy,
 at=(roc.outer south east),
 anchor=outer south west,
 xshift=7mm,
 height=\confmatlen,
 width=\confmatlen,
 title={\textit{LED Confidence}},
 title style={anchor=north, yshift=4mm, color=wong_blue},
 xlabel={Predicted Visibility},
 enlargelimits=false,
 xtick={0,1},
 xticklabels={No, Yes},
 ytick={0,1},
 yticklabels={No, Yes},
 yticklabel style={rotate=90},
 xtick style={draw=none},
 ytick style={draw=none},
 ticklabel style={font=\footnotesize},
 xlabel style={font=\small},
 ylabel style={font=\small},
 colormap name=allwhite,
 point meta min=000,
 point meta max=100,
]%
\addplot[matrix plot, point meta=explicit, nodes near coords={\pgfmathprintnumber{\pgfplotspointmeta}\%}, every node near coord/.style={anchor=center,color=black},font={\footnotesize}] coordinates {
 (0,0) [93] (1,0) [7]

 (0,1) [10] (1,1) [90]
};
\end{axis}%
\node at (cm_entropy.north west) [rotate=90,font=\small,anchor=center,xshift=4mm,yshift=9mm] {Robot Visibility};%
\begin{axis}[%
 name=cm_pos,
 at=(cm_entropy.above north west),
 anchor=south west,
 yshift=2mm,
 height=\confmatlen,
 width=\confmatlen,
 title={$\hat{P}_\text{ms-norm}$},
 title style={anchor=north, yshift=4mm, color=wong_pink},
 enlargelimits=false,
 xmajorticks=false,
 ytick={0,1},
 yticklabels={No, Yes},
 yticklabel style={rotate=90},
 xtick style={draw=none},
 ytick style={draw=none},
 ticklabel style={font=\footnotesize},
 xlabel style={font=\small},
 ylabel style={font=\small},
 colormap name=allwhite,
 point meta min=000,
 point meta max=100,
]%
\addplot[matrix plot, point meta=explicit, nodes near coords={\pgfmathprintnumber{\pgfplotspointmeta}\%}, every node near coord/.style={anchor=center,color=black},font={\footnotesize}] coordinates {
 (0,0) [37] (1,0) [63]

 (0,1) [10] (1,1) [90]
};
\end{axis}%
\end{tikzpicture}%
    \caption{Receiver Operator Characteristic (ROC) curves for the robot detection methods presented in Section~\ref{sec:results:table:detection} on the laboratory testing set.}
    \label{fig:enter-label}
\end{figure}

\end{document}